\pdfoutput=1
\RequirePackage{fix-cm}
\documentclass[sigconf,screen=true]{acmart}
\usepackage{subfig}
\usepackage{graphicx}

\usepackage[utf8]{inputenc}
\usepackage[T1]{fontenc}

\setcopyright{rightsretained}
\copyrightyear{2020}
\acmYear{2020}

\copyrightyear{2020}
\acmYear{2020}
\setcopyright{rightsretained}

\acmConference[IVA '20]{IVA '20: Proceedings of the 20th ACM International Conference on Intelligent Virtual Agents}{October 20--22, 2020}{Virtual Event, Scotland, UK}
\acmBooktitle{IVA '20: Proceedings of the 20th ACM International Conference on Intelligent Virtual Agents (IVA '20), October 20--22, 2020, Virtual Event, Scotland, UK}
\acmDOI{10.1145/3383652.3423874}
\acmISBN{978-1-4503-7586-3/20/09}

\begin{document}

\title{Generating coherent spontaneous speech and gesture from text}

\author{Simon Alexanderson}
\authornote{All authors are at KTH Royal Institute of Technology, Stockholm, Sweden.}
\affiliation{}
\email{simonal@kth.se}

\author{Éva Székely}
\authornotemark[1]
\affiliation{}
\email{szekely@kth.se}

\author{Gustav Eje Henter}
\authornotemark[1]
\affiliation{}
\email{ghe@kth.se}

\author{Taras Kucherenko}
\authornotemark[1]
\affiliation{}
\email{tarask@kth.se}

\author{Jonas Beskow}
\authornotemark[1]
\affiliation{}
\email{beskow@kth.se}



\begin{abstract}
Embodied human communication encompasses both verbal (speech) and non-verbal information (e.g., gesture and head movements). Recent advances in machine learning have substantially improved the technologies for generating synthetic versions of both of these types of data: On the speech side, text-to-speech systems are now able to generate highly convincing, spontaneous-sounding speech using unscripted speech audio as the source material. On the motion side, probabilistic motion-generation methods can now synthesise vivid and lifelike speech-driven 3D gesticulation. In this paper, we put these two state-of-the-art technologies together in a coherent fashion for the first time. Concretely, we demonstrate a proof-of-concept system trained on a single-speaker audio and motion-capture dataset, that is able to generate both speech and full-body gestures together from text input.
In contrast to previous approaches for joint speech-and-gesture generation, we generate full-body gestures from speech synthesis trained on recordings of spontaneous speech from the same person as the motion-capture data. 
We illustrate our results by visualising gesture spaces and text-speech-gesture alignments, and through a \href{https://simonalexanderson.github.io/IVA2020}{demonstration video}.
\end{abstract}



\settopmatter{printfolios=true} 
\maketitle

\section{Introduction}
\label{sec:intro}

Recent years have witnessed the rise of autonomous agents like digital assistants, embodied conversational agents and social robots.
While some of these communicate only through speech, others also exhibit non-verbal behaviours like gestures, head, and facial movements. Currently, both of these modalities are far from natural and are often criticised for appearing stiff and ``robotic''.
This is not surprising, given that text-to-speech (TTS) is typically trained on corpora of speech read aloud, while gestures often comprise concatenated snippets of ``canned'' motion. This stands in stark contrast to the smooth and spontaneous behaviours associated with human-human communication.
Recently, independent research in both TTS and gesture synthesis has shown that deep learning can remedy some of these issues. 
In speech synthesis, Tacotron \cite{shen2018natural} has ushered in a quantum leap in speech prosody modelling, 
greatly increasing the vividness of synthesised speech. 
For motion synthesis, emerging probabilistic modelling techniques such as MoGlow \cite{henter2019moglow} have been shown to generate highly convincing gestures from speech input \cite{alexanderson2020style}.


In this paper, we demonstrate 
how existing state-of-the-art, data-driven TTS and gesture generation systems can be put together to synthesise both speech and gesture at the same time simply from text input. To the best of our knowledge, our system is the first data-driven system capable of synthesising coherent verbal and non-verbal behavior from the same actor using text as only input. This constitutes a first step towards a unified model that generates both modalities simultaneously.
Our key innovations are:
\begin{enumerate}
\item We generate hand and full-body gestures from \textit{synthetic speech} (as opposed to recordings of natural speech).
\item We drive gesture-generation using TTS trained on recordings of \textit{spontaneous speech}. This is important since traditional TTS uses recordings of texts being read aloud, which is not ideal since humans do not gesture when reading.
\item Speech and gesture-generation is trained on data from the \textit{same person}, so the gesture behaviour matches the speech.
\end{enumerate}

\section{Background}
Recent advances in TTS based on sequence-to-sequence models with neural attention \cite{shen2018natural} have succeeded in generating realistic-sounding speech with vivid and lifelike prosody from text alone.
These techniques have also shown unprecedented robustness to challenging data, with \cite{szekely2019spontaneous} recently demonstrating that Tacotron 2 is able to create highly convincing TTS from recordings of spontaneous speech with imprecise automatic transcriptions.
The new ability to move away from TTS corpora of read speech and produce convincingly spontaneous-sounding synthetic speech is particularly relevant to our interest in generating coherent co-speech gestures, since the vast majority of human speech is spontaneous and part of a dialogue. 
In addition, novel methods have been proposed to automatically control conversational phenomena such as breathing and disfluencies \cite{Szekely2020breathing}. 
We leverage these breakthroughs for the speech synthesis in this paper.


Synthesis of non-verbal human behaviour has gradually moved from rule-based systems towards data-driven approaches. Current deep neural models typically use 
recurrent neural networks 
to predict a sequence of continuous gestures from speech represented either as audio \cite{hasegawa2018evaluation, kucherenko2019analyzing} or text \cite{yoon2018robots}. Here we employ the latest progress in probabilistic gesture models using normalising flows \cite{alexanderson2020style}, which has been shown to generate highly diverse and believable results. To our knowledge, no previous work has been able to simultaneously output gesture and speech from text input, nor investigated the effects of driving gesture generation from TTS output.

\section{System description}
\label{sec:system}
We construct our multimodal speech-and-gesture-generation system by chaining together two essential pieces, namely a spontaneous text-to-speech synthesiser and a speech-audio-driven gesture generator.
In this section we describe these two components. 
Both components were trained on the Trinity Speech-Gesture Dataset \cite{ferstl2018investigating}, which comprises 31 episodes (244 minutes) of audio and full-body motion capture 
of a male actor speaking spontaneously (with disfluencies and fillers) on different topics while shifting stance and moving freely through the capture area.
We held out the last episode as test data.


\subsection{Spontaneous speech synthesis}
\label{ssec:tts}
Our data annotation and speech synthesis pipeline followed the approach described by \cite{szekely2019spontaneous}.
To detect breath events and segment the corpus, we used a simplified version of the speaker-dependent breath detector proposed in \cite{szekely2019casting}. 
This enabled us to segment the audio recordings into 3,487 automatically-detected \emph{breath groups} (speech segments delineated by consecutive breaths).
To create a synthesis corpus, these breath groups were paired up into bigrams, which allows synthesising longer utterances and for taking context into account across utterances.
The voice was then built based on the Tacotron 2 spectrogram-prediction framework \cite{shen2018natural}, using public resources for transcription, TTS, pre-training, phonetisation, and waveform synthesis, following \cite{szekely2019spontaneous}. 
Breath events and hesitations (uh and um) were given a unique phone symbol.

\subsection{Gesture generation}
\label{ssec:moglow}
For full-body gesture generation we used a technique called MoGlow \cite{henter2019moglow}, adapted to gestures as described in \cite{alexanderson2020style}.
The model is autoregressive, and learns the distribution for the next motion pose given previous poses and a context window of speech features from the corresponding speech audio.
New motion is generated by randomly sampling a sequence of poses from this next-step distribution.
This produces different (but plausible) gestures every time, even if the speech input is the same.

Setup and data processing followed \cite{alexanderson2020style}.
The aligned gestures and natural speech acoustics were downsampled to 20 frames (poses and mel-spectra) per second.
Poses were parameterised using joint rotations,
complemented by the forwards, sideways, and angular displacement of the root, to enable the system to generate small steps and changes in stance based on the input.
The data was also augmented by lateral mirroring. We used the same network size as in \cite{alexanderson2020style} and trained the model for 80,000 steps using a data dropout rate of 0.4 and a learning rate decaying from $\textrm{lr}_{\textrm{max}}=2\cdot10^{-3}$ to $5\cdot10^{-4}$.
At synthesis time, 20 fps mel-spectrum features from either natural or synthetic speech waveforms were fed into the system to sample matching synthetic pose sequences.


\section{Evaluation and discussion}
\label{sec:evaluation}
We demonstrate the efficacy of our
system
in several ways.
First, we prepared a video of the system presenting itself by synthesising a character from text input 
(\href{https://simonalexanderson.github.io/IVA2020}{simonalexanderson.github.io/IVA2020}).
Then, we compare the gesture spaces induced by natural and synthetic speech input to the gesture generator. Finally, we study the alignment between speech input and sampled output gestures.
The stimuli for these evaluations were three utterances selected from the held-out data, 12 to 16 s long, each comprising three sequential breath groups.
We transcribed the words and original breath locations in the selected utterances and fed these to the TTS.
%
%
\begin{figure}[t]
\centering
\vspace{-4mm}
\subfloat[Gestures from natural speech.]{\includegraphics[width=0.45\linewidth]{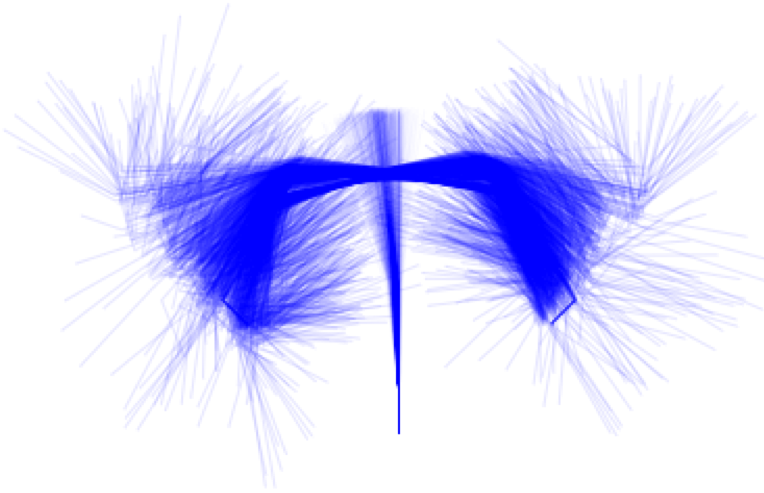}}
\hspace{5mm}
\subfloat[Gestures from synthetic speech.]{\includegraphics[width=0.46\linewidth]{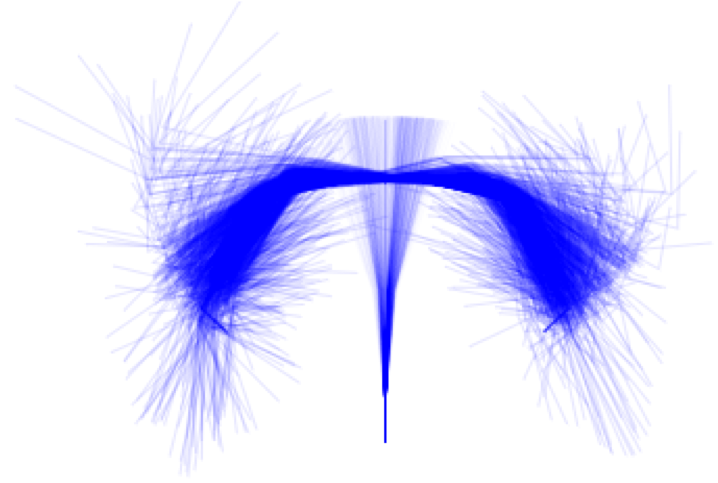}}\\
\vspace{-3mm}
\subfloat[Training data before mirroring.]{\includegraphics[width=0.46\linewidth]{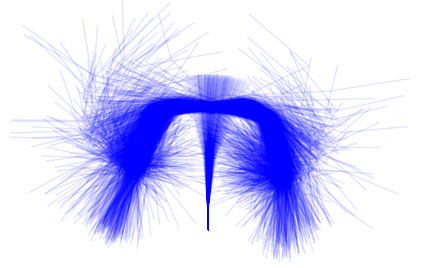}}
\hspace{5mm}
\subfloat[Mirrored training data.]{\includegraphics[width=0.46\linewidth]{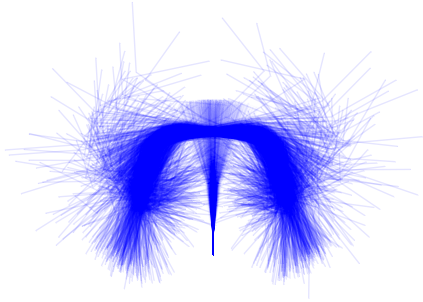}}
\caption{Gesture-space visualisation for model output from natural and synthetic speech, and for training data.}
\label{fig:gesture_space}
\vspace{-1.5em}
\end{figure}

\subsection{Gesture-space comparison}
Many data-driven gesture-synthesis methods suffer from regression to the mean pose, making synthetic gestures more constrained and less vivid than the natural training gestures, as illustrated in \cite{ferstl2019multi}.
To check this, we visualise the gesture spaces of model-synthesised and natural gestures
by overlaying a hip-centered upper-body skeleton of the gesticulating avatar extracted every 8th frame for 15 random samples from each of the three selected utterances.
The result is shown in Fig.~\ref{fig:gesture_space}.
We see that the gesture spaces for gestures synthesised from the model are virtually the same as the natural gesture space of the training data.
This contrasts against other gesture-synthesis methods such as those visualised in \cite[Fig.~1]{ferstl2019multi}.
Crucially, gestures generated from synthetic speech acoustics (Fig.~\ref{fig:gesture_space}b) essentially fill the same space as gestures synthesised from natural speech input (Fig.~\ref{fig:gesture_space}a).
This suggests that our approach generalises well to synthetic speech and still occupies the entire gesticulation space that a human naturally would use (as also seen in our demonstration video).
Making motion statistics, e.g., the spacial extent of gestures, match natural gesticulation is also important since these statistics correlate with the perception of personality traits and emotional states
\cite{castillo2019we,koppensteiner2010motion}.
%
%
\begin{figure*}[!t]
\centering
\includegraphics[width=1.00\textwidth]{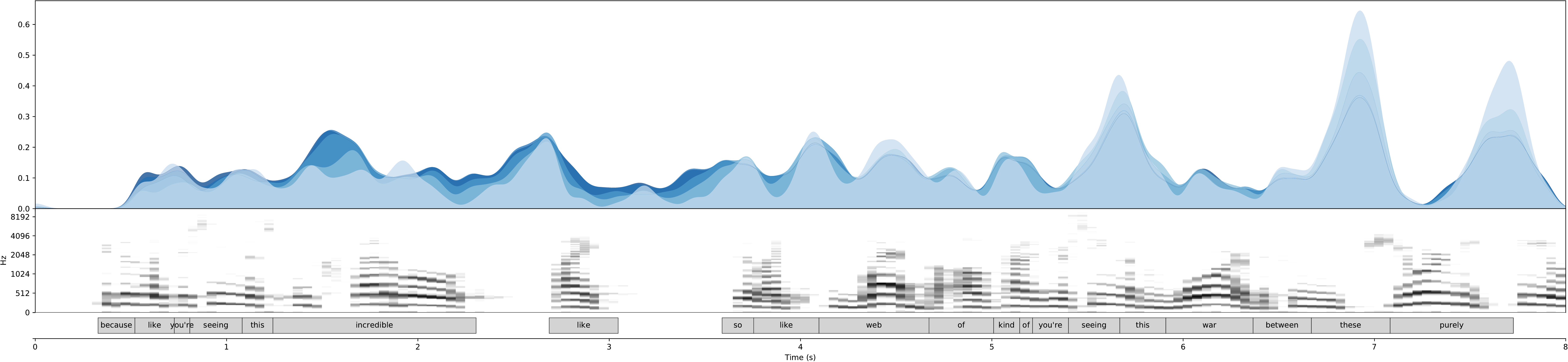}\\
\includegraphics[width=1.00\textwidth]{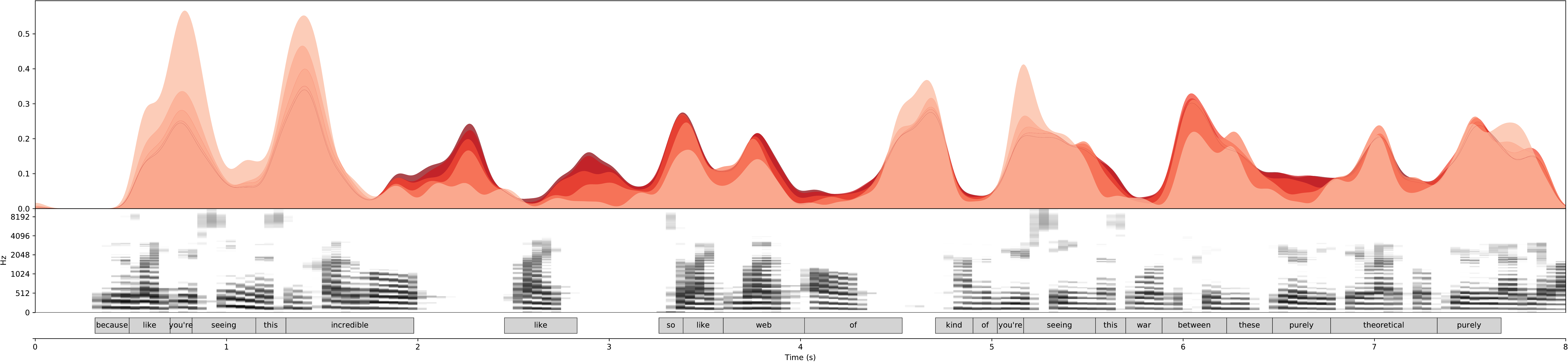}
\caption{Velocity peak distribution. The colour plots illustrate how the top $N$ ($2 \leq N \leq 12$) hand-velocity peaks of the synthesised motion are distributed for 300 pose sequences sampled for one natural (top/blue) and one synthetic (bottom/red) speech input utterance. Darker shades correspond to higher $N$. 
Below each distribution plot is the mel spectrogram (the system input features) and the timings of the spoken words.}
\label{fig:velprofile}
\vspace{-1.5em}
\end{figure*}

\subsection{Alignment between speech and gesture}
A key aspect of our approach, where the gesture synthesis is driven by synthetic speech, is that we can generate co-speech gestures that align with speech timings.
Since the approach is probabilistic, we can get a good impression of this alignment using only a single utterance but many random draws, as seen in Fig.~\ref{fig:velprofile}, which focuses on the first 8 s of one of the selected stimulus utterances.
The figure uses kernel density estimation to visualise the distribution of the top peaks in hand velocity -- taken as indicators of peak gesticulation intensity -- across 300 randomly sampled pose sequences for the same input audio.
There is clear variation between realisations, just as human gestures never are the same every time.
However, the gesticulation intensity is not uniformly random, but exhibits a consistent structure, with pronounced peaks near emphasised speech segments and low gesture activity near long pauses.
A deterministic gesture-generation method would require us to analyse hundreds of different utterances to reliably see this kind of structure.
The differences in velocity distribution between gestures driven by natural speech audio (blue, top pane) and gestures driven by synthesised speech (red, lower pane) in Fig.~\ref{fig:velprofile} are not unexpected, since the intonation and emphasis of the text-derived synthetic speech is not the same as in the natural speech.
\section{Conclusion and future work}
\label{sec:conclusion}
We have demonstrated multimodal synthesis of both speech and full-body co-speech gestures from the same input text, by chaining together state-of-the-art text-to-speech and speech-to-gesture systems.
Using the latest developments in spontaneous speech synthesis enables gestures to be generated from convincing spontaneous-style TTS audio.
By using speech audio and motion capture sourced from a single person, we generate speech and gesture behaviours that match a real human individual.
Future work involves tighter integration between gesture and speech synthesis, by learning a single, unified model that generates both modalities simultaneously.

\bibliographystyle{ACM-Reference-Format}
\bibliography{refs}


\end{document}